\pdfoutput=1

\documentclass[11pt]{article}
\usepackage[final]{acl}
\usepackage{times}
\usepackage{latexsym}
\usepackage[T1]{fontenc}
\usepackage[utf8]{inputenc}
\usepackage{microtype}
\usepackage{inconsolata}
\usepackage{graphicx}
\usepackage{amsmath}
\usepackage{algorithm}
\usepackage{algpseudocode}
\usepackage{xcolor}
\usepackage{colortbl}
\usepackage{booktabs}

\title{Language-Aware Token Boosting:\\LLM Language Confusion Reduction Without Tuning}

\author{%
  Trapoom Ukarapol$^{\mathsection\ddagger}$, Pakhapoom Sarapat$^\mathsection$, Nut Chukamphaeng$^{\mathsection \dagger}\thanks{This work was carried out during the author's tenure at SCB DataX.}$
   \\
   $^\mathsection$SCB DataX, $^\ddagger$Tsinghua University, $^\dagger$SCBX \\
  \texttt{\{trapoom.ukarapol, pakhapoom.sarapat\}@data-x.ai, nut.c@scbx.com}\\
}

\begin{document}
\maketitle
\begin{abstract}
Large language models (LLMs) sometimes exhibit language confusion when generating non-English text. Existing approaches typically rely on fine-tuning to mitigate this issue. In contrast, we propose a tuning-free paradigm for reducing language confusion. Within this paradigm, we introduce two methods: Language-Aware Token Boosting (LATB), which applies targeted perturbations to tokens associated with the desired language, and Adaptive Language-Aware Token Boosting (Adaptive-LATB), which dynamically adjusts these perturbations based on the model’s confidence in the intended language. Experiments demonstrate that our methods effectively improve multilingual alignment by reducing language confusion, while maintain the summarization quality without requiring any additional fine-tuning. Our code is publicly available.\footnote{\url{https://github.com/scbdatax/genai-datax-language-aware-token-boosting}}.

\end{abstract}

\section{Introduction}

Large Language Models (LLMs) have shown impressive performance, but their English-centric development limits their effectiveness for non-English users \cite{hadi2024large, hadi2023survey}. Recent efforts \cite{xue2021mt5massivelymultilingualpretrained, workshop2023bloom176bparameteropenaccessmultilingual, wei2023polylmopensourcepolyglot} aim to enhance multilingual capabilities, though English-centric models still underperform in low-resource languages \cite{qin2024multilinguallargelanguagemodel, openai2024gpt4technicalreport}. One of the key issues is language confusion \cite{devine2024tagengomultilingualchatdataset}, where models fail to consistently generate the desired language, particularly in non-English contexts \cite{marchisio2024understandingmitigatinglanguageconfusion}. Techniques to mitigate this include temperature lowering, few-shot prompting, and fine-tuning \cite{marchisio2024understandingmitigatinglanguageconfusion}, but these come with limitations such as reduced responses diversity \cite{Agarwal_2024, renze2024effectsamplingtemperatureproblem} or increased computational costs.

\begin{figure}[t]
  \includegraphics[width=\columnwidth]{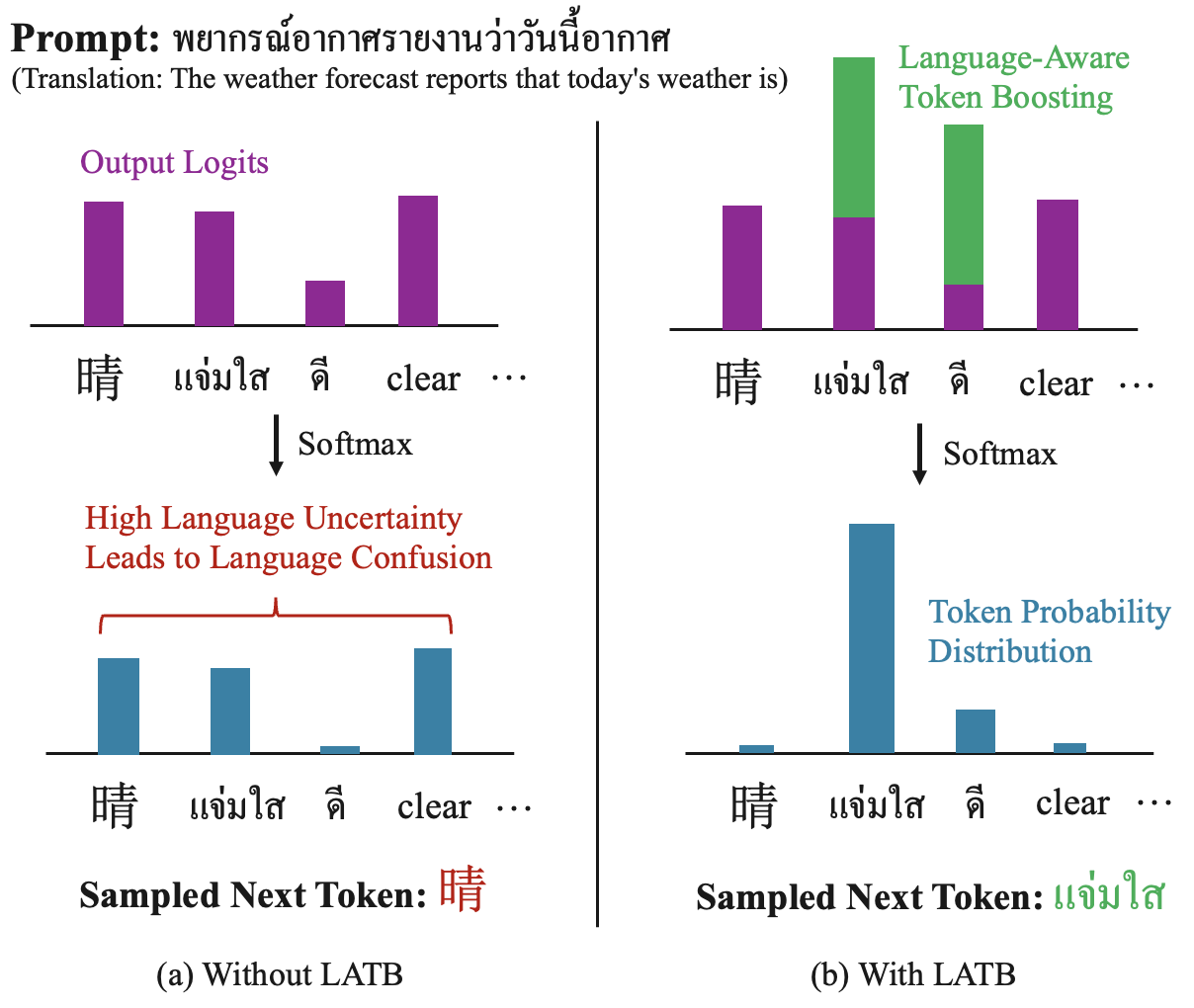}
  \caption{Language-Aware Token Boosting (LATB) enhances target language generation confidence by selectively boosting target language tokens.}
  \label{fig:main}
\end{figure}

We propose a novel tuning-free paradigm for multilingual alignment, using perturbations directly on the logits. This approach eliminates the need for fine-tuning and aligns the model's outputs with the desired language, incurring minimal additional computational costs during inference. We introduce two methods within this paradigm: \textbf{Language-Aware Token Boosting (LATB)}, which applies language-specific token perturbations, and \textbf{Adaptive Language-Aware Token Boosting (Adaptive-LATB)}, which adapts perturbations by introducing perturbations selectively—only when the LLM exhibits uncertainty in generating one language over another.

We evaluate our methods on the XLSUM multilingual summarization benchmark \cite{hasan2021xlsumlargescalemultilingualabstractive} across eight languages. Both LATB and Adaptive-LATB effectively reduce language confusion and maintain summarization performance compared to their respective base models and the multilingual-tuned model.

In summary, our contributions are as follows:  
\begin{enumerate}  
    \item We propose a novel tuning-free multilingual alignment paradigm based on logits perturbation, introducing two methods: LATB and Adaptive-LATB.  
    \item We evaluate our methods on the XLSUM benchmark, showing reduced language confusion and maintain summarization quality.  
    \item We analyze the effect of logit perturbations and show that adding a constant value to the logits does not change the relative probability ratios among the selected tokens.
\end{enumerate}

\section{Related Work}

\paragraph{Multilingual Large Language Models.} Multilingual Large Language Models (MLLMs) are designed to process multiple languages simultaneously. The approaches for developing and optimizing these models can be broadly categorized into two main types: parameter-tuning alignment (PTA) and parameter-frozen alignment (PFA) \cite{qin2024multilinguallargelanguagemodel}. The PTA approach involves tuning the model’s parameters to enable multilingual capabilities. This tuning can occur at various stages, including pretraining \cite{xue2021mt5massivelymultilingualpretrained, chowdhery2022palmscalinglanguagemodeling, workshop2023bloom176bparameteropenaccessmultilingual, jiang2023mistral7b, jiang2024mixtralexperts}, supervised fine-tuning (SFT) \cite{chung2022scalinginstructionfinetunedlanguagemodels, muennighoff2023crosslingualgeneralizationmultitaskfinetuning, devine2024tagengomultilingualchatdataset, pipatanakul2023typhoonthailargelanguage}, reinforcement learning with human feedback (RLHF) \cite{lai2023okapiinstructiontunedlargelanguage, touvron2023llama2openfoundation, glm2024chatglmfamilylargelanguage, bai2023qwentechnicalreport}, and downstream task fine-tuning \cite{lepikhin2020gshardscalinggiantmodels, rosenbaum2022linguistlanguagemodelinstruction}. In contrast, PFA methods do not require parameter tuning for multilingual alignment. Instead, they primarily rely on prompting techniques \cite{abdelali2024larabenchbenchmarkingarabicai, winata2023decadesprogresscodeswitchingresearch, lu2024chainofdictionarypromptingelicitstranslation, puduppully2023decomposedpromptingmachinetranslation} and retrieval-augmented alignment \cite{he2023exploringhumanliketranslationstrategy, zhang-etal-2023-leveraging, conia2023increasingcoverageprecisiontextual}.
Our proposed method falls within the PFA category. To the best of our knowledge, our study is the first to introduce a new taxonomy for logits perturbation-based multilingual alignment.
\paragraph{Language Confusion.} Language confusion refers to the inconsistent ability of LLMs to generate responses in a target language. This phenomenon has been observed across a range of NLP tasks, including machine translation \cite{vu2022overcomingcatastrophicforgettingzeroshot, li2023doeszeroshotcrosslingualgeneration}, summarization \cite{wang2023understandingtranslationesecrosslingualsummarization, yu-etal-2022-translate}, question answering \cite{holtermann2024evaluatingelementarymultilingualcapabilities}, and even within the reasoning traces of reasoning language models (RLMs) \cite{wang2025languagemixingreasoninglanguage, tam2025languagemattersmultilingualinput}. While this issue has been systematically studied with various proposed methods mitigating it \cite{marchisio2024understandingmitigatinglanguageconfusion}, our study introduces a novel and cost-effective approach to mitigate language confusion using token perturbation methods.

\section{Approach}

\subsection{Token Language Identification}

We identify tokens to boost based on the target language using a Unicode filtering method following \cite{wen-yi-mimno-2023-hyperpolyglot}. Specifically, a token is considered valid if all its characters belong to the Unicode set defined for the target language. We also include numbers, special characters, and the end of sentence tokens in the desired set.

\subsection{Perturbation Vector}

We construct a perturbation vector, $\mathbf{p}$, based on the set of desired token indices $I$. Each element corresponding to an index in $I$ is assigned a perturbation value $\alpha \ge 0$, as defined in Equation \ref{eq:p_vec}.

\begin{equation}
    \label{eq:p_vec}
    \mathbf{p}_i = 
    \begin{cases} 
    \alpha & \text{if } i \in I, \\
    0 & \text{otherwise.}
    \end{cases} 
\end{equation}

\subsection{Logits Perturbation Methods}

In this study, we explore two variants of the Logits Perturbation Method: LATB and Adaptive-LATB.

\subsubsection{Language-Aware Token Boosting (LATB)}

We introduce perturbations to the logits by adding a perturbation value $\alpha$ to the selected logits to align them with the desired language. The method is detailed in Algorithm \ref{alg:LATB}.

\begin{algorithm}
\caption{Vanilla LATB}\label{alg:LATB}
\begin{algorithmic}
\State $\textbf{logits} \gets LLM(x)$
\State $\textbf{logits}' \gets \textbf{logits} + \mathbf{p}$ \Comment{Logits Perturbation}
\State $\mathbf{y'} \gets \text{Softmax}(\textbf{logits}')$
\end{algorithmic}
\end{algorithm}

\subsubsection{Adaptive Language-Aware Token Boosting (Adaptive-LATB)}

Adding logits in the vanilla LATB may suppress the ability to express tokens in another language when necessary. In contrast, the Adaptive LATB perturbs logits only when the LLM is not confident about the language it intends to express. The confidence difference threshold, controlled by the hyperparameter $\beta$ ($0 \leq \beta \leq 1$), determines the model's confidence difference threshold in one language over another. This design enables the model to switch languages when it is highly confident. The details of the Adaptive LATB algorithm are provided in Algorithm \ref{alg:adaptive_LATB}.

\begin{algorithm}
\caption{Adaptive LATB}\label{alg:adaptive_LATB}
\begin{algorithmic}
\State $\textbf{logits} \gets LLM(x)$
\State $\mathbf{y} \gets \text{Softmax}(\textbf{logits})$
\State $a \gets \text{max}(\{ y_i \mid y_i \in \mathbf{y} \text{ and } i \in I \})$
\State $b \gets \text{max}(\{ y_i \mid y_i \in \mathbf{y} \text{ and } i \not\in I \})$
\If{$| a-b | < \beta$}
\State $\textbf{logits}' \gets \textbf{logits} + \mathbf{p}$ \Comment{Logits Perturbation}
\State $\mathbf{y'} \gets \text{Softmax}(\textbf{logits}')$
\Else
\State $\mathbf{y'} \gets \mathbf{y}$
\EndIf
\end{algorithmic}
\end{algorithm}

\section{Evaluation Metrics} 

We evaluate the model based on two key aspects: \textit{Language Confusion}, which measures the model's misalignment with the target language, and \textit{Performance}, which assesses the quality of the generated summaries.

\subsection{Language Confusion Metrics} 

We evaluate language confusion at three distinct
levels to capture both fine-grained and overall effects:
token-level, line-level, and response-level
language confusion.

\paragraph{Token-level Language Confusion.} 

We determine each token's language based on its Unicode and calculate token-level misalignment rates for each response. These rates are then averaged across all responses to report the final metric.

\paragraph{Line-level Language Confusion.} 

We segment
each response by line and utilize an off-the-shelf
language identification (LID) tool, FastText \cite{joulin2016bag, joulin2016fasttext}, to determine the language of each line.
We calculate the average language misalignment
per response and report the overall average across
all responses.

\paragraph{Response-level Language Confusion.} 

We input the entire response into the FastText \cite{joulin2016bag, joulin2016fasttext} language identification and calculate the average language misalignment across all responses, reporting this as the final metric.

\subsection{Performance Metrics} 

We assess summarization performance using three widely adopted metrics: ROUGE-1, ROUGE-2, and ROUGE-L \cite{lin-2004-rouge}. These metrics evaluate the overlap of unigrams, bigrams, and longest common subsequences, respectively, between the generated summaries and the reference summaries.

\section{Experiments}

\begin{table*}[htbp]
\caption{Language confusion across different methods evaluated on eight languages, reported as Token-level/Line-level/Response-level language confusion in percentage.}
\label{tab:res_lc}
\resizebox{\textwidth}{!}{
\begin{tabular}{@{}lccccc@{}}
\toprule
  & Llama3 8B-I & Llama3 8B-I (Strict Prompt) & Suzume 8B-Multilingual & Llama3 8B-I + LATB (\textit{Ours}) & Llama3 8B-I + Adaptive LATB (\textit{Ours}) \\ \midrule
\multicolumn{6}{c}{\cellcolor[HTML]{C0C0C0}High Resource Languages (HRL)}  \\
ru & 80.66/93.83/92.50 & 5.02/4.10/2.90 & 3.04/2.30/2.10 & \textbf{0.28}/0.44/\textbf{0.10} & 0.48/\textbf{0.38}/\textbf{0.10} \\
zh & 85.92/98.90/98.90 & 14.17/9.69/9.10 & 7.56/0.89/0.90 & \textbf{4.78}/\textbf{0.00}/\textbf{0.00} & 5.37/0.10/\textbf{0.00} \\
ja & 85.10/98.83/98.31 & 10.15/9.29/4.16 & 5.96/0.73/0.67 & \textbf{3.51}/0.70/\textbf{0.11} & 4.05/\textbf{0.16}/\textbf{0.11} \\
fr & 0.24/47.14/40.6 & 0.26/0.39/\textbf{0.20} & 0.31/0.37/0.30 & \textbf{0.11}/0.35/\textbf{0.20} & 0.18/\textbf{0.25}/0.30 \\
\toprule
\multicolumn{6}{c}{\cellcolor[HTML]{C0C0C0}Medium Resource Languages (MRL)} \\
ko & 85.46/99.60/99.63 & 16.72/30.79/27.27 & 8.28/11.74/12.36 & \textbf{3.45}/\textbf{9.98}/\textbf{10.36} & 4.56/10.12/11.45 \\
th & 86.03/99.80/99.39 & 3.67/9.80/2.30 & 2.16/1.16/0.84 & 0.43/0.18/\textbf{0.00} & \textbf{0.38}/\textbf{0.00}/\textbf{0.00} \\
hi & 86.18/99.05/98.50 & 1.67/8.59/0.40 & 2.77/3.36/2.50 & \textbf{0.23}/\textbf{0.74}/0.10 & 0.26/0.89/\textbf{0.00} \\
ar & 86.21/99.27/98.30 & 9.98/11.94/5.60 & 5.63/2.95/2.60 & \textbf{0.37}/0.28/\textbf{0.00} & 0.54/\textbf{0.22}/\textbf{0.00} \\
\bottomrule
\end{tabular}
}
\end{table*}

\begin{table*}[htbp]
\caption{Summarization performance across different methods evaluated on eight languages, reported as ROUGE-1/ROUGE-2/ROUGE-L in percentage.}
\label{tab:res_rouge}
\resizebox{\textwidth}{!}{
\begin{tabular}{@{}lccccc@{}}
\toprule
  & Llama3 8B-I & Llama3 8B-I (Strict Prompt) & Suzume 8B-Multilingual & Llama3 8B-I + LATB (\textit{Ours}) & Llama3 8B-I + Adaptive LATB (\textit{Ours}) \\ \midrule
\multicolumn{6}{c}{\cellcolor[HTML]{C0C0C0}High Resource Languages  (HRL)}  \\
ru & 4.89/0.96/4.18 & 20.44/9.26/13.41 & 19.35/8.32/12.42 & 20.83/\textbf{9.46}/\textbf{13.60} & \textbf{21.00}/9.42/13.58
\\
zh & 0.80/0.32/0.69 & 19.41/8.99/13.73 & 19.31/8.59/13.38 & \textbf{20.70}/\textbf{9.44}/\textbf{14.64} & 20.55/9.28/14.52 \\
ja & 26.42/12.53/16.84 & 26.48/12.43/16.97 & 26.13/11.73/16.55 & 27.54/\textbf{12.95}/17.70 & \textbf{27.89}/12.92/\textbf{17.89} \\
fr & 14.71/6.09/10.49 & 19.98/8.90/13.71 & 18.56/7.89/12.47 & \textbf{20.13}/\textbf{9.05}/\textbf{13.74} & 19.97/8.89/13.59 \\
\toprule
\multicolumn{6}{c}{\cellcolor[HTML]{C0C0C0}Medium Resource Languages (MRL)} \\
ko & 2.27/0.24/2.12 & 14.66/6.14/10.16 & 15.30/6.13/10.47 & 16.41/6.78/11.38 & \textbf{16.88}/\textbf{7.03}/\textbf{11.67} \\
th & 1.79/0.45/1.51 & 29.24/13.99/15.62 & 28.99/13.29/15.14 & 29.77/14.07/15.79 & \textbf{30.97}/\textbf{14.74}/\textbf{16.41} \\
hi & 0.93/0.36/0.73 & 29.83/16.41/19.03 & 27.71/14.78/17.52 & 29.68/16.41/19.00 & \textbf{29.77}/\textbf{16.41}/\textbf{19.05} \\
ar & 1.54/0.19/1.42 & 19.22/7.46/11.66 & 19.60/7.09/11.69 & \textbf{20.44}/\textbf{8.02}/\textbf{12.45} & 19.79/7.62/11.84 \\
\bottomrule
\end{tabular}
}
\end{table*}

\paragraph{Experimental Setup.}
We compare Llama3 8B Instruct \cite{lai2023chatgptenglishcomprehensiveevaluation} under several configurations: (i) a \emph{normal prompt} in the target generation language without any language constraints in the prompt, (ii) a \emph{strict prompt} that explicitly specifies the target generation language, (iii) a multilingual supervised fine-tuned variant of Llama3 8B Instruct, namely Suzume 8B Multilingual \cite{devine2024tagengomultilingualchatdataset}, and (iv) our proposed tuning-free methods, Language-Aware Token Boosting (LATB) and Adaptive-LATB. Llama3 8B Instruct serves as the base English-centric model, while Suzume 8B Multilingual provides a strong multilingual fine-tuning baseline for comparison.

For evaluation, we adopt the multilingual abstractive summarization benchmark XLSUM \cite{hasan2021xlsumlargescalemultilingualabstractive}. This dataset is well suited to our study as it requires models to generate long-form outputs, enabling systematic quantitative evaluation. We consider eight languages spanning different resource levels, including four High-Resource Languages (HRL): Russian (ru), Simplified Chinese (zh), Japanese (ja), and French (fr), and four Medium-Resource Languages (MRL): Korean (ko), Thai (th), Hindi (hi), and Arabic (ar). The classification of language resource levels follows \cite{lai2023chatgptenglishcomprehensiveevaluation}. For each language, we randomly sample up to 1,000 test instances for evaluation.

\paragraph{Summarization Quality Results.} 
 The results reported in Table~\ref{tab:res_rouge} demonstrate that our methods preserve generation quality and yield marginal improvements, with gains increasing as the degree of language confusion rises prior to applying LATB, as detailed in Appendix~\ref{apd:perf_improvement}.

\section{Analysis}

\paragraph{Effect of Logit Perturbations on Boosted Token Probabilities.}

Let $I$ denote the set of boosted tokens. Let $y_i$ and $y'_i$ be the probabilities of the $i$-th token before and after logit perturbation, respectively. Denote the logit of the $i$-th token by $v_i$, and let $\alpha$ be the perturbation magnitude applied to all tokens in $I$. We analyze the probability mass function over the boosted tokens before and after token boosting. Consider any two tokens $m, n \in I$.

\begin{align}
\frac{y'_m}{y'_n}
&= \frac{\left(e^{v_m + \alpha}\right) / \left(\sum_{i \notin I} e^{v_i} + \sum_{i \in I} e^{v_i + \alpha}\right)}
         {\left(e^{v_n + \alpha}\right) / \left(\sum_{i \notin I} e^{v_i} + \sum_{i \in I} e^{v_i + \alpha}\right)} \\
         &= \frac{e^{v_m}}{e^{v_n}} = \frac{e^{v_m} / \sum_i e^{v_i}}{e^{v_n} / \sum_i e^{v_i}} = \frac{y_m}{y_n}.
\label{eq:prove}
\end{align}

Therefore, the logit perturbation preserves the relative probability ratios between any pair of boosted tokens.

\paragraph{Vanilla vs. Adaptive LATB.}

Both methods deliver comparable performance. However, Vanilla LATB requires an optimal hyperparameter search to produce non-target language output when needed while accurately generating results in the target language. In contrast, Adaptive LATB is less sensitive to hyperparameters and supports non-target language output as required.

\paragraph{Impact on Inference Speed.}

We evaluated inference throughput of Llama3 8B Instruct \cite{lai2023chatgptenglishcomprehensiveevaluation} with vLLM on an A100 GPU. The base model achieves 1145.8 tokens/s, while vanilla LATB achieves 1189.5 tokens/s effectively identical throughput within measurement noise. This is expected, as LATB only applies a lightweight logit bias without additional branching or search. Adaptive-LATB reaches 838 tokens/s, reflecting its extra computation for per-step maximum probabilities difference detection. Although this introduces overhead, the throughput remains within a practical range for deployment.

\section{Conclusion}

This paper introduces a novel approach to multilingual alignment for English-centric language models through token perturbation techniques. We proposed the Language-Aware Token Boosting (LATB) and its adaptive variant, Adaptive-LATB. Extensive experiments demonstrate that our methods significantly reduce language confusion compared to base model and outperform its multilingual fine-tuned model. This highlights the efficiency and practicality of our approach for enhancing multilingual language model capabilities.

\section*{Limitations and Future Work}

Our work shows promising results but has several limitations. First, the methods struggle with aligning LLMs to untrained or out-of-vocabulary (OOV) tokens. Second, reliance on Unicode-based language identification is less effective for languages with significant overlap with Latin scripts. Finally, hyperparameter tuning is needed to balance language confusion and multilingual expression. Future work could improve OOV token handling, develop better token-based language identification techniques, and design language-agnostic hyperparameter selection methods.

\bibliography{custom}

\appendix

\section{Experiment Details} 
We generate responses using the Llama3 8B Instruct model \cite{grattafiori2024llama3herdmodels} on eight different languages from the XLSUM dataset \cite{hasan2021xlsumlargescalemultilingualabstractive}. The prompts utilized for this experiment are detailed in Appendix \ref{apd:prompt}. All responses are generated with the sampling parameters set to a temperature of 1.0 and a top-$p$ value of 1.0. For LATB, the perturbation value $\alpha$ is set to 5. For Adaptive LATB, the perturbation value is set to $\alpha = 1000$, and the confidence difference threshold is set to $\beta = 0.8$.

\section{Prompt Templates}
\label{apd:prompt}

\begin{figure}[ht]
  \includegraphics[width=\columnwidth]{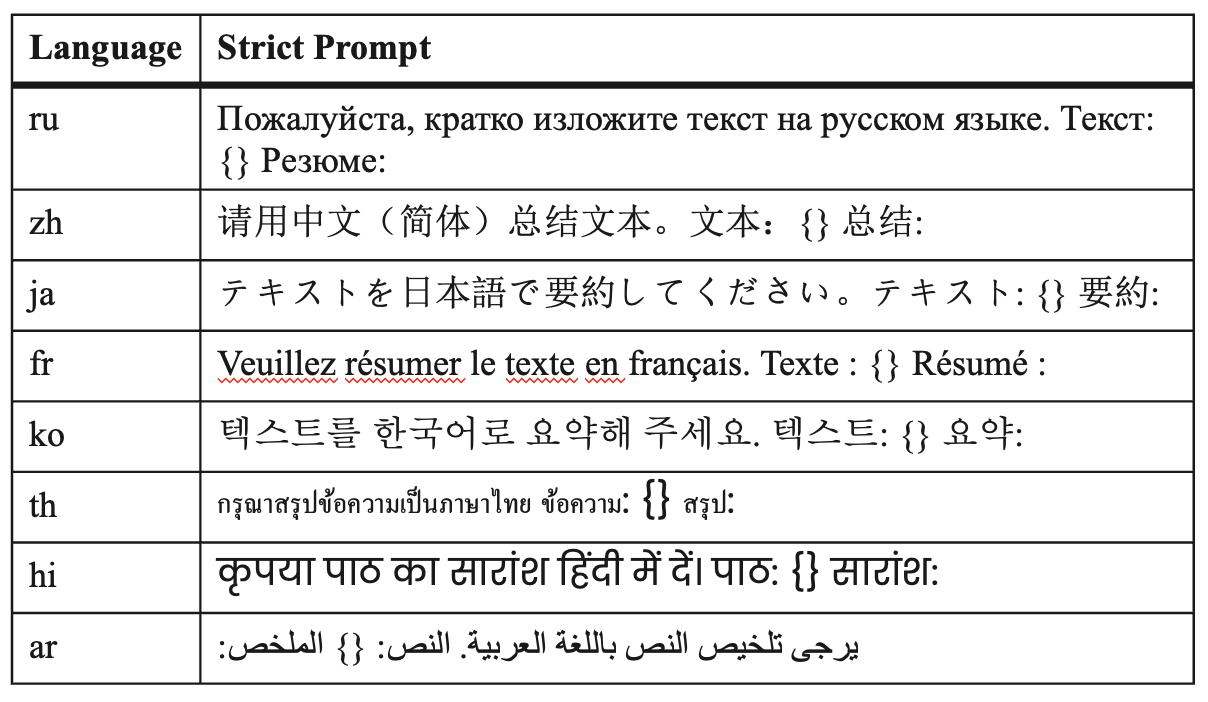}
  \caption{Strict prompt templates used in the experiment}
  \label{fig:strict_prompt_template}
\end{figure}

\begin{figure}[ht]
  \includegraphics[width=\columnwidth]{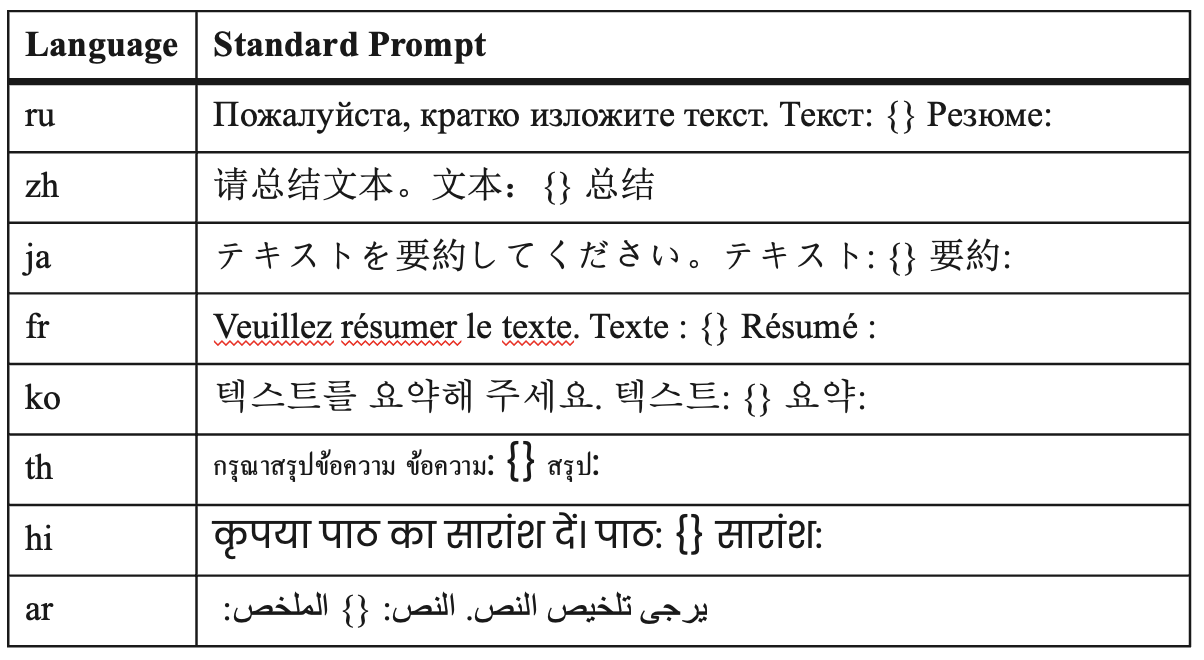}
  \caption{Standard prompt templates used in the experiment}
  \label{fig:standard_prompt_template}
\end{figure}

We design language-specific prompt templates to ensure consistency and adaptability across different languages during text generation. Each template provides a structured format where \{\} is replaced by the input text to summarize. The strict prompt templates include instructions to ensure the model generates output in the target language, whereas the standard prompt templates do not. The standard and strict prompt templates are shown in Figures~\ref{fig:standard_prompt_template} and~\ref{fig:strict_prompt_template}, respectively.

\begin{figure*}[ht]
  \includegraphics[width=0.48\linewidth]{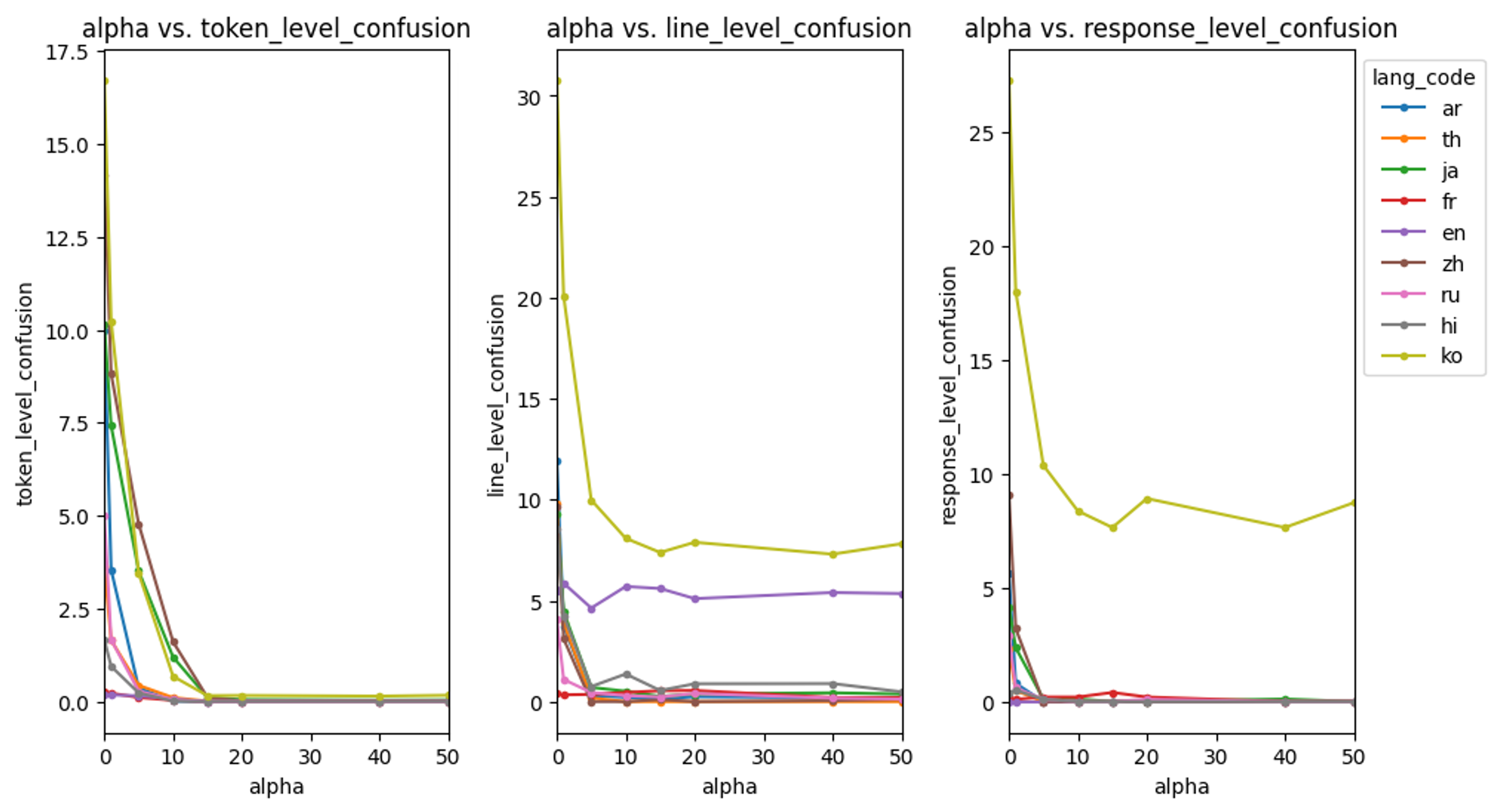} \hfill
  \includegraphics[width=0.48\linewidth]{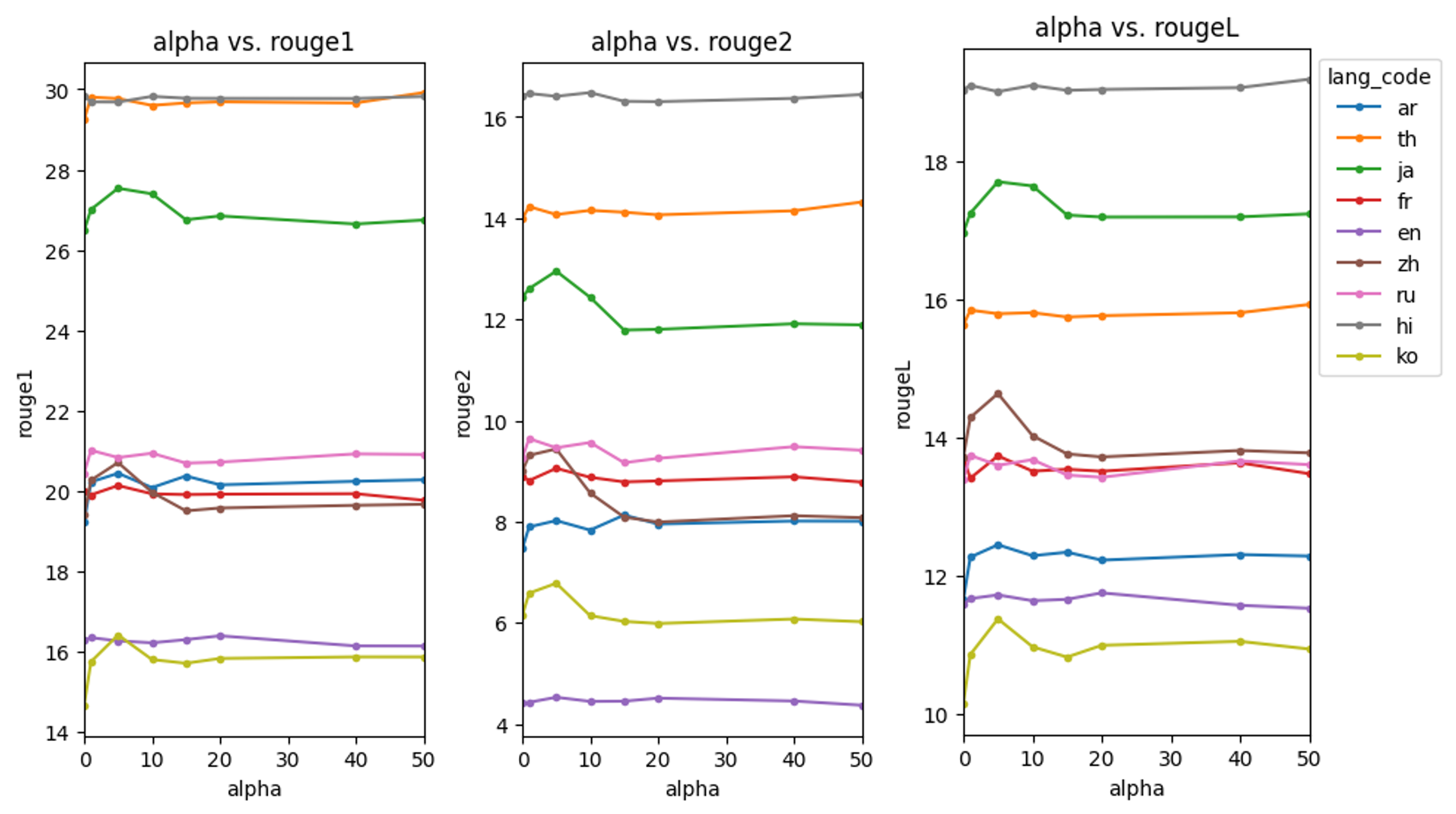}
  \caption{Impact of the Perturbation Value $\alpha$ on Language Confusion and Performance in LATB}
  \label{fig:alpha_varying}
\end{figure*}

\begin{figure*}[ht]
  \includegraphics[width=0.48\linewidth]{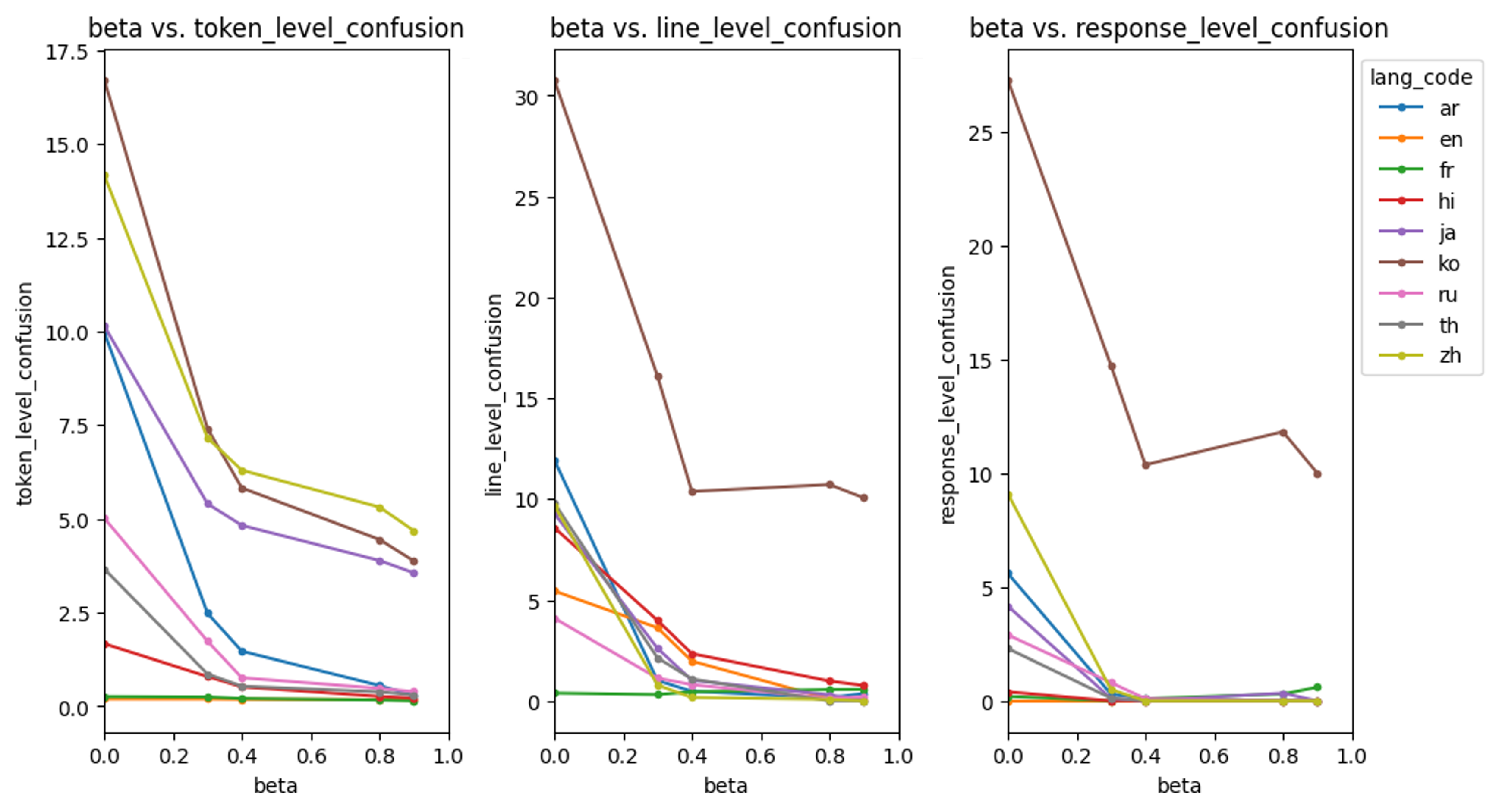} \hfill
  \includegraphics[width=0.48\linewidth]{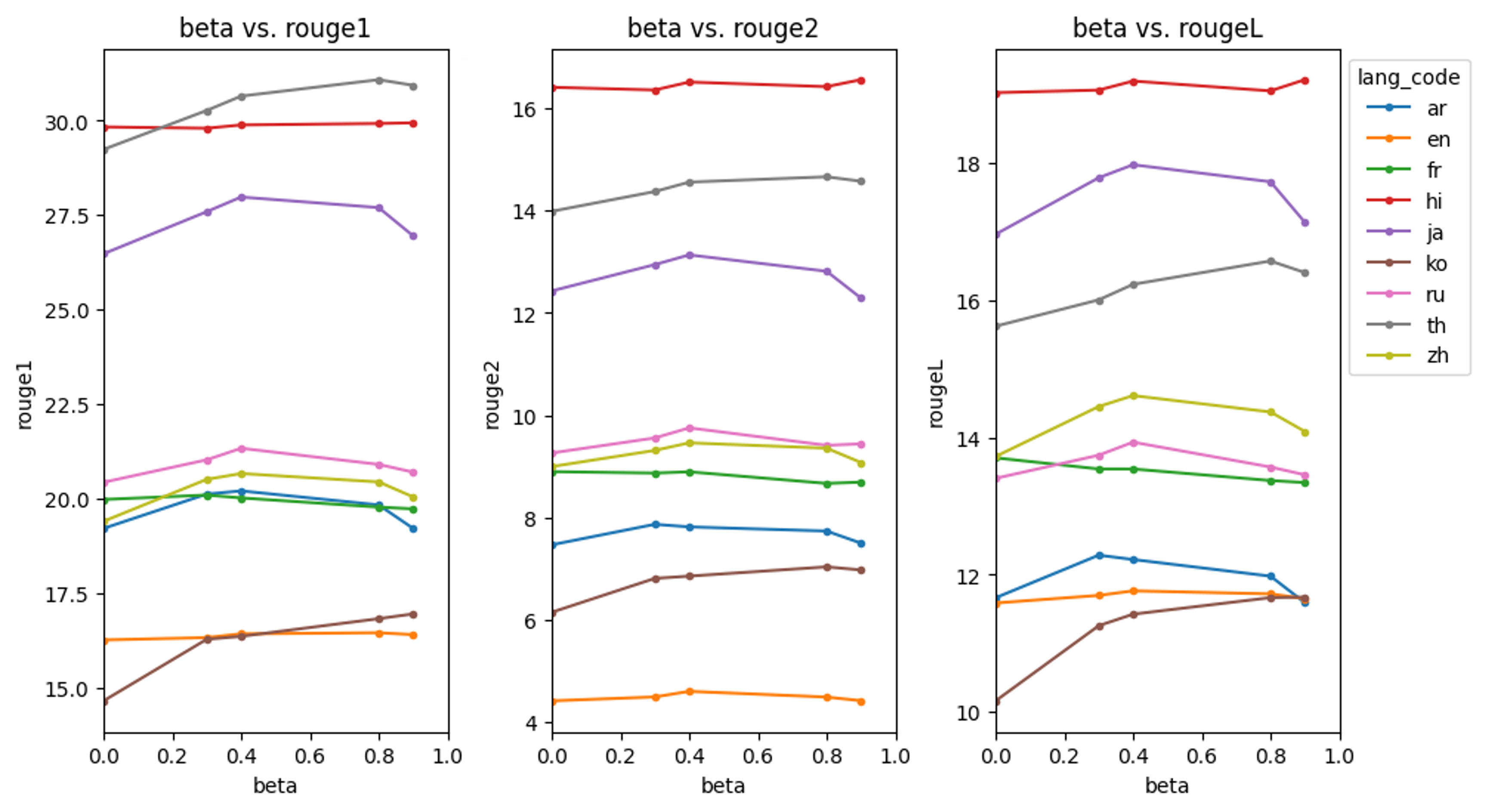}
  \caption{Impact of the Confidence Difference Threshold $\beta$ on Language Confusion and Performance in Adaptive-LATB with $\alpha$ fixed at 1000}
  \label{fig:beta_varying}
\end{figure*}

\section{Impact of Hyperparameters}
\label{apd:impact_hyper}

We analyze the impact of hyperparameters in both LATB and Adaptive-LATB on language confusion and summarization quality. All responses across experiments were generated with a temperature of 1.0 and a top-$p$ value of 1.0 to ensure consistent sampling.

\paragraph{LATB.}
In LATB, the perturbation parameter $\alpha$ plays a critical role in controlling language confusion. We varied $\alpha$ from 0 to 50 and report the results in Figure~\ref{fig:alpha_varying}. As $\alpha$ increases, language confusion is progressively reduced, leading to an initial improvement in ROUGE scores. At an intermediate value of $\alpha$, the model achieves an effective balance: it can express technical terms in English while minimizing language confusion at both the line and response levels. Beyond this optimal point, further increasing $\alpha$ overly suppresses tokens from non-target languages, which degrades summarization quality and results in a decline in ROUGE scores.

\paragraph{Adaptive-LATB.}
For Adaptive-LATB, we study the effect of the confidence difference threshold $\beta$ while fixing the perturbation value $\alpha$ to 1000. We varied $\beta$ from 0 to 0.9, with results shown in Figure~\ref{fig:beta_varying}. Increasing $\beta$ generally reduces language confusion and yields slight improvements in ROUGE scores up to an inflection point. However, excessively large $\beta$ values restrict the model’s ability to generate necessary non-target language tokens, leading to a mild performance drop. This trend highlights a similar trade-off between reducing language confusion and preserving generation flexibility.

\begin{table*}[htbp]
\caption{Qwen3 4B-I Language confusion and Summarization performance across different methods evaluated on eight languages reported as Response-level language confusion/ROUGE-L in percentage.}
\label{tab:res_qwen}
\resizebox{\textwidth}{!}{
\begin{tabular}{@{}lccccc@{}}
\toprule
  & Qwen3 4B-I & Qwen3 4B-I (Strict Prompt) & Qwen3 4B-I + LATB (\textit{Ours}) & Qwen3 4B-I + Adaptive LATB (\textit{Ours}) \\ \midrule
\multicolumn{5}{c}{\cellcolor[HTML]{C0C0C0}High Resource Languages (HRL)}  \\
ru & 0.00/13.45 & 0.00/13.16 & 0.00/13.30 & 0.00/12.68 \\
zh & 0.00/12.42 & 0.00/12.58 & 0.00/12.49 & 0.00 /12.26\\
ja & 0.00/17.63 & 0.00/17.01 & 0.00/16.88 & 0.00/16.90 \\
fr & 0.00/13.33 & 0.00/12.99 & 0.00/13.02 & 0.00/12.74 \\
\toprule
\multicolumn{5}{c}{\cellcolor[HTML]{C0C0C0}Medium Resource Languages (MRL)} \\
ko & 0.00/12.34 & 0.00/11.78 & 0.00/11.51 & 0.00/11.55 \\
th & 0.00/18.64 & 0.00/17.77 & 0.00/17.60 & 0.00/17.13 \\
hi & 0.00/21.81 & 0.00/21.58 & 0.00/16.57 & 0.00/18.42 \\
ar & 0.00/11.67 & 0.00/10.94 & 0.00/11.14 & 0.00/10.06 \\
\bottomrule
\end{tabular}
}
\end{table*}

\begin{figure}[t]
  \includegraphics[width=\columnwidth]{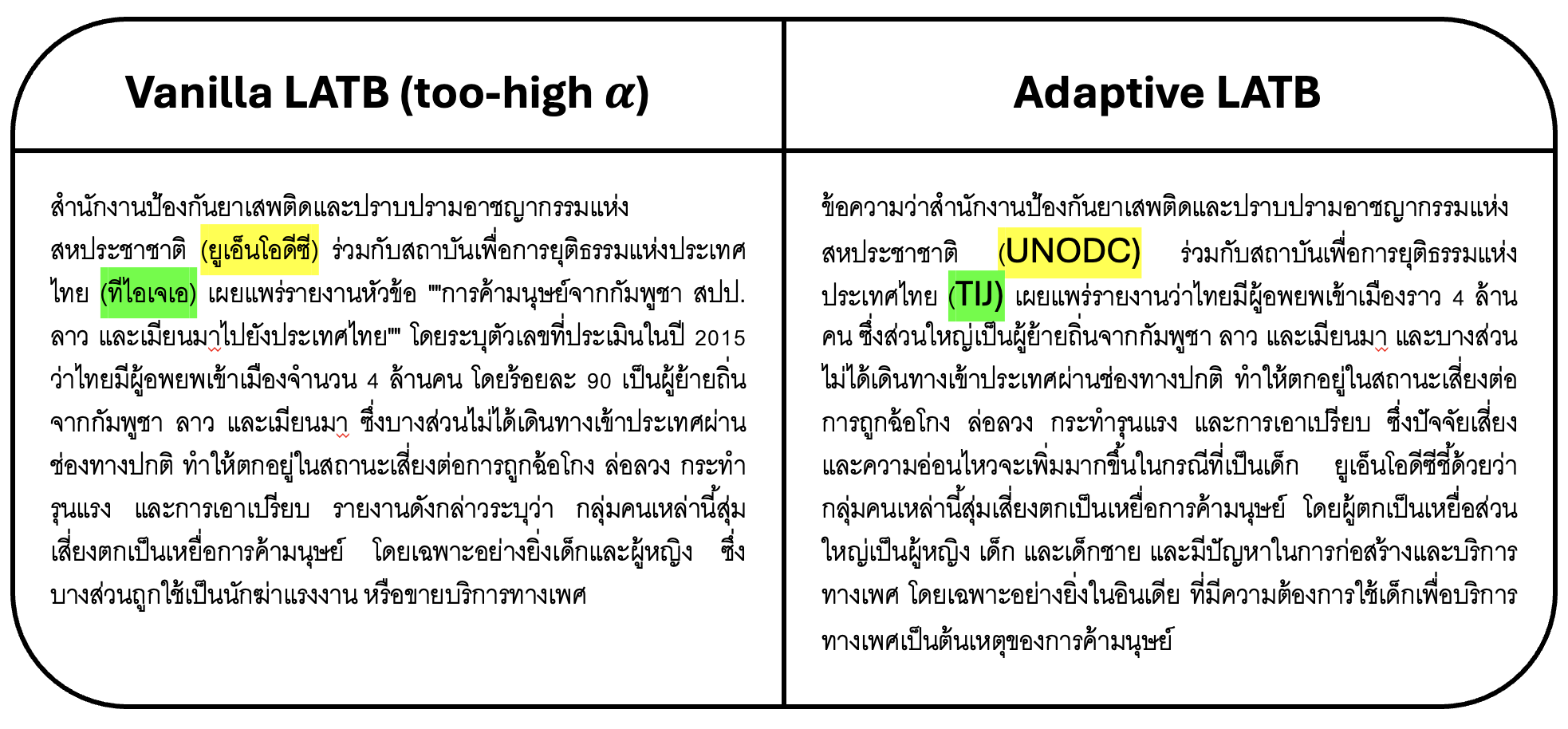}
  \caption{Output examples of Vanilla LATB with excessively high $\alpha$ and Adaptive LATB}
  \label{fig:vanilla-vs-adaptive}
\end{figure}

\section{Vanilla vs. Adaptive LATB Example}

As discussed, Vanilla LATB is sensitive to hyperparameters, which can lead to a constraint on single-target language generation when $\alpha$ is too high. In contrast, Adaptive LATB is less sensitive to hyperparameters. The example output is shown in Figure~\ref{fig:vanilla-vs-adaptive}. When $\alpha$ is too high, Vanilla LATB generates a Thai-dubbed version of English, whereas Adaptive LATB uses English directly, resulting in a more natural output.

\section{Additional Experiments on a Stronger Multilingual Model}
We extend our experiments to Qwen3-4B-I \cite{yang2025qwen3technicalreport} and measure response-level language confusion and summarization quality using the ROUGE-L metric. The base model already demonstrates strong language alignment with no observed errors, and our method preserves this behavior: it achieves similarly error-free language alignment.

Importantly, our method does not degrade task performance. As shown in Eq. \ref{eq:prove}, LATB preserves the relative probability ratios among tokens within the same language; therefore, intra-language token ranking and consequently generation quality remains unchanged. Consistent with this theoretical property, we observe no loss in ROUGE-L compared to the base model. The results are shown in Table \ref{tab:res_qwen}.

\section{Performance Improvements} 
\label{apd:perf_improvement}
\begin{figure}[ht]
  \includegraphics[width=\linewidth]{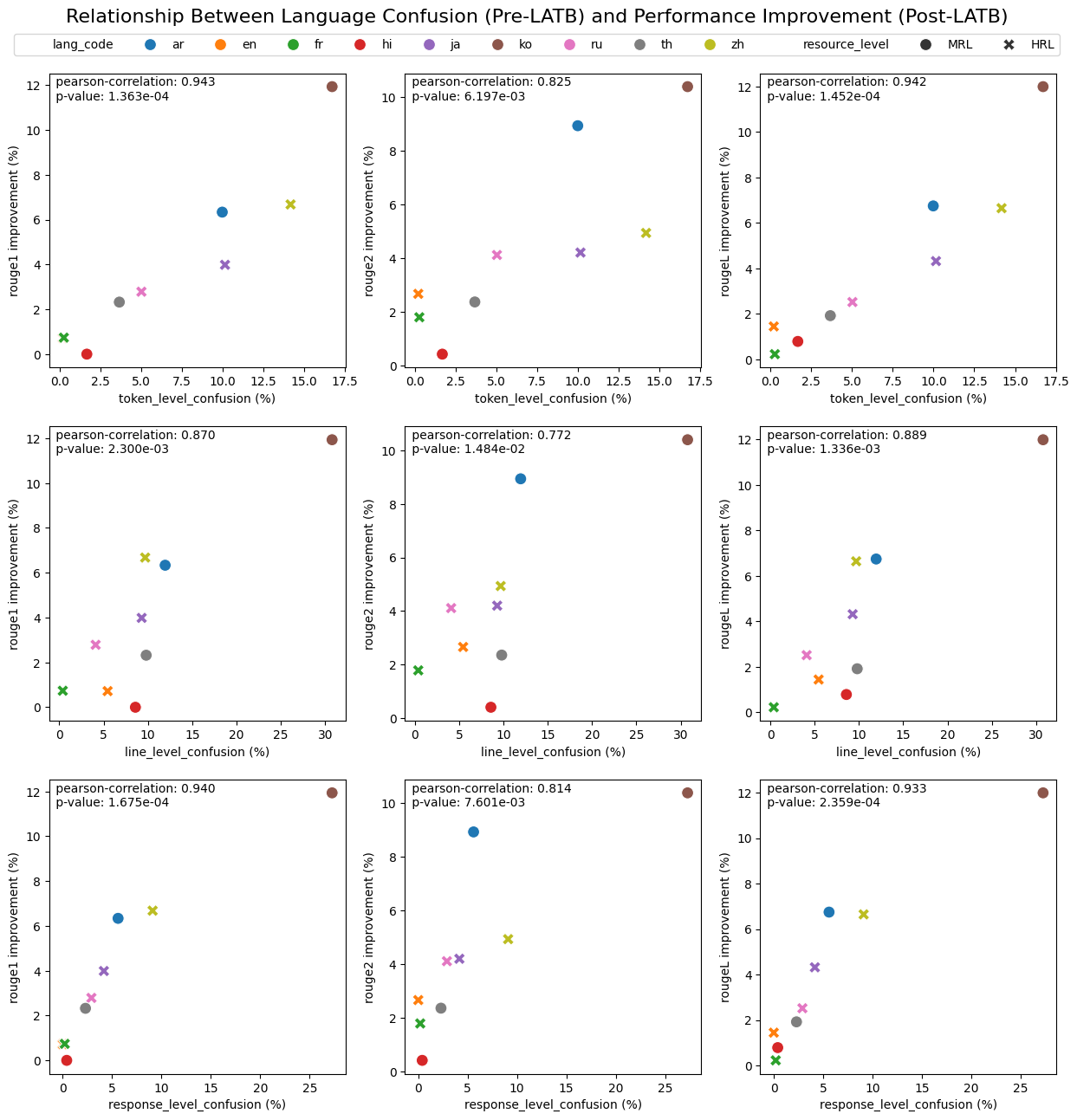}
  \caption{Performance improvements with LATB correlate strongly with language confusion levels}
  \label{fig:corr}
\end{figure}

Our analysis highlights a strong correlation between performance improvements from LATB and the degree of language confusion without LATB. This finding suggests that language confusion contributes to performance degradation. By incorporating LATB, we effectively mitigate this issue, leading to performance gains. The relationship is illustrated in Figure \ref{fig:corr}.

\end{document}